\documentclass[11pt,a4paper]{article}

\usepackage[utf8]{inputenc}
\usepackage{geometry}
\usepackage{amsmath,amssymb}
\usepackage{bm}
\usepackage{algorithm}
\usepackage{algpseudocode}
\usepackage{physics}
\usepackage[normalem]{ulem}
\usepackage{graphicx}
\usepackage{subfig}
\usepackage{float}
\usepackage{cite}
\usepackage{tabularx}
\usepackage{booktabs}
\usepackage{hyperref}
\usepackage{authblk}

\newcolumntype{C}{>{\centering\arraybackslash}X}

\title{Hybrid Quantum-Classical AI for Industrial Defect Classification in Welding Images}

\author[1,2,*]{Akshaya Srinivasan}
\author[1]{Xiaoyin Cheng}
\author[3]{Jianming Yi}
\author[1,*]{Alexander Geng}
\author[4]{Desislava Ivanova}
\author[5]{Andreas Weinmann}
\author[1,*]{Ali Moghiseh}

\affil[1]{Fraunhofer Institute for Industrial Mathematics ITWM, 67663 Kaiserslautern, Germany}
\affil[2]{RPTU Kaiserslautern-Landau, 67663 Kaiserslautern, Germany}
\affil[3]{MuVision UG, Gasstra{\ss}e 21, 67655 Kaiserslautern, Germany}
\affil[4]{Technical University of Sofia, 8 Kliment Ohridski Blvd, 1000 Sofia, Bulgaria}
\affil[5]{Algorithms for Computer Vision, Imaging and Data Analysis Lab, Technical University of Applied Sciences, W{\"u}rzburg-Schweinfurt, Ignaz-Sch{\"o}n-Str. 11, Schweinfurt, Germany}

\affil[*]{Correspondence: akshaya.srinivasan@itwm.fraunhofer.de; alexander.geng@itwm.fraunhofer.de; ali.moghiseh@itwm.fraunhofer.de}

\date{}

\begin{document}

\maketitle

\begin{abstract}
Hybrid quantum-classical machine learning offers a promising direction for advancing automated quality control in industrial settings. In this study, we investigate two hybrid quantum-classical approaches for classifying defects in aluminium TIG welding images and benchmarking their performance against a conventional deep learning model. A convolutional neural network is used to extract compact and informative feature vectors from weld images, effectively reducing the higher-dimensional pixel space to a lower-dimensional feature space. Our first quantum approach encodes these features into quantum states using a parameterized quantum feature map composed of rotation and entangling gates. We compute a quantum kernel matrix from the inner products of these states, defining a linear system in a higher-dimensional Hilbert space corresponding to the support vector machine (SVM) optimization problem and solving it using a Variational Quantum Linear Solver (VQLS). We also examine the effect of the quantum kernel condition number on classification performance. In our second method, we apply angle encoding to the extracted features in a variational quantum circuit and use a classical optimizer for model training. Both quantum models are tested on binary and multiclass classification tasks and the performance is compared with the classical CNN model. Our results show that while the CNN model demonstrates robust performance, hybrid quantum-classical models perform competitively. This highlights the potential of hybrid quantum-classical approaches for near-term real-world applications in industrial defect detection and quality assurance.
\end{abstract}

\vspace{1em}

\section{Introduction} \label{sec:intro}

\begin{figure}[H]
    \centering
    \includegraphics[width=\textwidth]{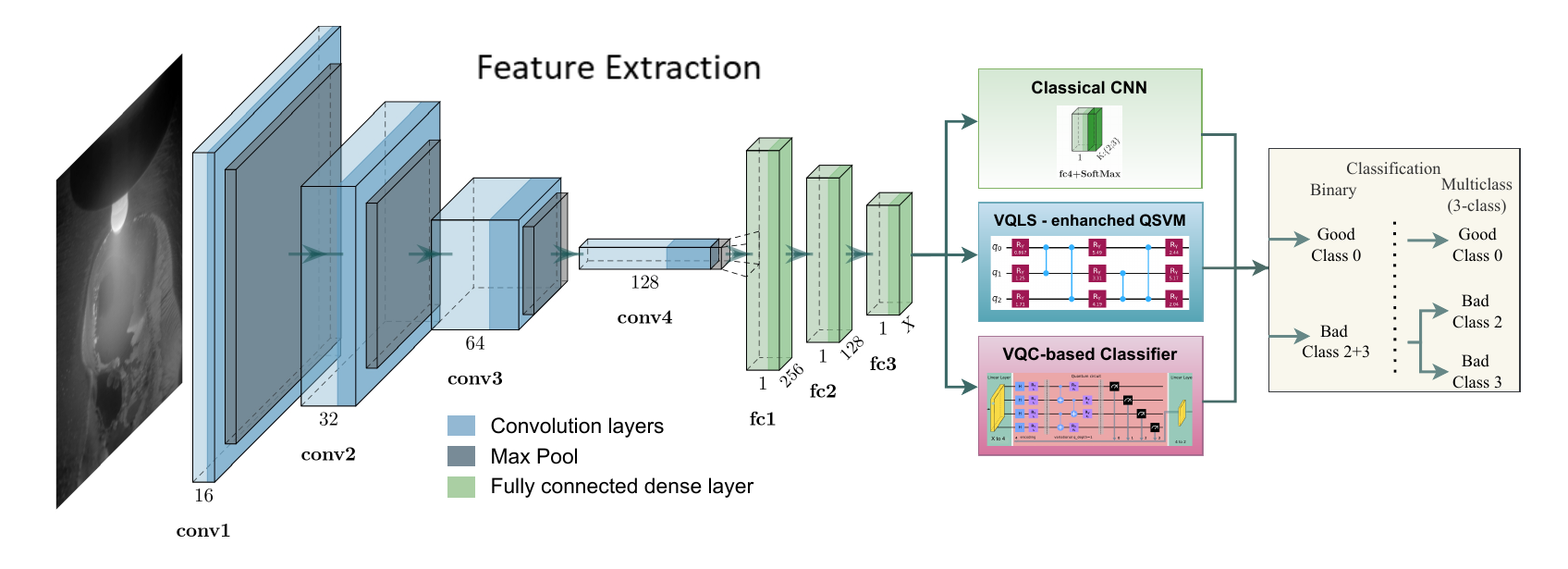}
    \caption{Full architecture of the study, showing CNN-based feature extraction followed by three classification approaches: classical CNN, VQLS-enhanced QSVM, and VQC-based classifier.\label{fig:overview}}
\end{figure}

Automated quality control is a cornerstone of modern industrial manufacturing, enabling early defect detection, process optimization, and assurance of product quality. In sectors such as the automotive industry, aerospace, and heavy machinery, welding is a critical joining process in which errors can compromise structural safety. Convolutional neural networks (CNNs) have achieved notable success in image-based classification tasks, including weld quality assessment \cite{zhang2019weld,yaping2019research}. However, these models demand significant computational resources, particularly as data dimensionality and model complexity increase. In parallel with the evolution of classical machine learning, quantum computing has emerged as a novel computational paradigm, exploiting principles of superposition and entanglement to process information in fundamentally different ways. While fault-tolerant quantum computing remains a long-term goal, the current noisy intermediate-scale quantum (NISQ) era \cite{preskill2018quantum} has enabled the development of hybrid quantum-classical models that combine quantum circuits with classical optimization and pre-/post-processing. These hybrid approaches have shown promise in domains such as machine learning, finance, and computer vision tasks (see, e.g., \cite{alam2021quantum,biamonte2017quantum,schuld2021machine,geng_hybrid_2025,geng_hybrid_2022,geng_improved_2023,geng_application_2024,SrinivasanGengMacalusoKieferEmmanouilidisMoghiseh+2025}). Quantum machine learning (QML), especially in its hybrid quantum-classical form, offers new possibilities for extracting information from data with potentially fewer parameters or enhanced generalization capabilities. In this work, we investigate two hybrid quantum-classical approaches for the classification of weld defects in aluminium TIG (tungsten inert gas) welding images. Figure~\ref{fig:overview} gives an overview of our work.

We begin by extracting features from welding images using a CNN architecture adapted from Model-1 described in the study \cite{bacioiu_automated_2019}. Unlike generic pretrained models such as ResNet18 \cite{paszke_pytorch_2019,russakovsky_imagenet_2015}, we selected Model-1 because it was specifically designed and tested for weld image classification. Out of the 13 architectures evaluated in that study, Model-1 showed superior performance, making it well-suited for our task. We trained this model end-to-end on our dataset to generate feature vectors for subsequent classification. This part serves as a feature extractor, transforming higher-dimensional pixel space into a lower-dimensional feature space. These extracted features form the input for two distinct quantum models, too.

Our first approach employs a quantum kernel method, introduced by Yi et al. in \cite{yi_variational_2023}: Feature vectors are embedded into quantum states using a parameterized feature map composed of rotation and entangling gates. A quantum kernel matrix is constructed via inner products of these states, effectively mapping the data into a higher-dimensional Hilbert space. This kernel is integrated into a support vector machine (SVM) formulation, where the optimization problem is solved using the Variational Quantum Linear Solver (VQLS), a hybrid algorithm designed to approximate the solution to linear systems on quantum hardware \cite{bravo2023variational, rebentrost_quantum_2014}. Additionally, we investigate the condition number of the quantum kernel matrix and its influence on classification performance and numerical stability. The model is tested on a range of classification tasks, including binary and multiclass scenarios, to measure how well it performs.

In the second approach, we use a variational quantum circuit (VQC) where classical features are angle-encoded into quantum gates. A parameterized quantum circuit is trained using a classical optimizer to minimize a loss function defined by the classification task. This model is evaluated on both binary and multiclass classification problems to assess its expressivity and performance under varying task complexities.

By comparing these hybrid methods with a classical CNN baseline, this study aims to assess the feasibility of quantum models in real-world industrial applications. Through this comparative analysis, we seek to better understand the potential and limitations of current quantum machine learning approaches in the context of weld defect image classification.

\section{Algorithms} \label{sec:algo}

\subsection{Feature Extraction}

Due to the current limitations of NISQ-era quantum hardware, particularly in terms of qubit count, gate fidelity, and circuit depth, processing high-resolution image data directly on quantum processors is not yet feasible. Encoding an entire image of even moderate resolution requires a large number of qubits. Downsizing the input too aggressively risks discarding vital spatial and structural information. To overcome this challenge, we adopt a hybrid classical-quantum approach in which the classical component performs dimensionality reduction to make the data suitable for quantum processing. A convolutional neural network (CNN) is used to extract meaningful features from the raw images, compressing them into a lower-dimensional, information-rich vector. This preserves essential image characteristics while keeping the input dimensions within the feasible range for current quantum devices.

\begin{figure}[H]
    \centering
    \includegraphics[width=0.7\textwidth]{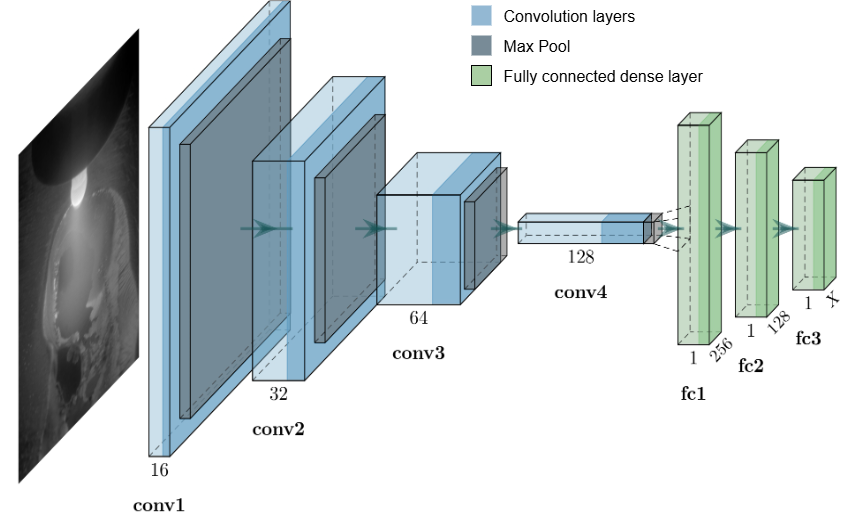}
    \caption{CNN architecture for Feature Extraction. The last fully connected layer outputs a feature vector $X \in \{7,15,31,63,127\}$.\label{fig:feature-extract}}
\end{figure}

As illustrated in Figure~\ref{fig:feature-extract}, feature vectors are extracted from a fully connected (dense) layer of the CNN. These vectors serve as inputs for the subsequent quantum processing steps. We systematically evaluated multiple feature sizes specifically, vectors of length 7, 15, 31, 63, and 127 to determine the optimal dimensionality for each quantum pipeline. The goal was to balance quantum circuit depth and fidelity against the classification performance of the model. This intermediate feature representation serves a dual purpose: it reduces the input dimensionality for quantum encoding and captures the most relevant information from the input data.
By integrating classical convolutional layers with quantum models, we design an interface that transforms complex, higher-dimensional image data into simpler, lower-dimensional information that quantum circuits can process effectively.

\subsection{Classical Convolutional Neural Network}

\begin{figure}[H]
    \centering
    \includegraphics[width=0.7\textwidth]{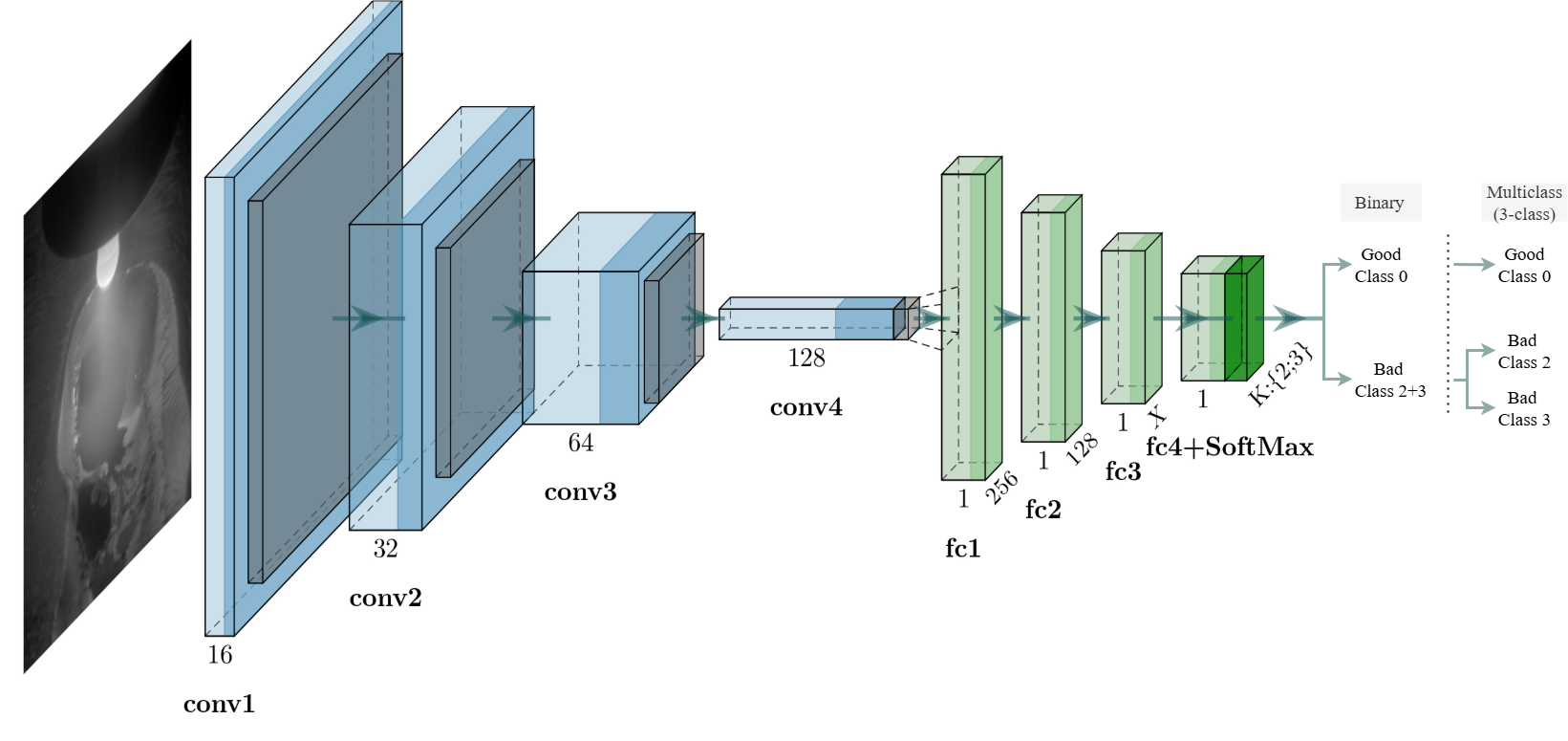}
    \caption{Classical CNN architecture for classification.\label{fig:cnn}}
\end{figure}

As a baseline for performance benchmarking, we implement a conventional convolutional neural network (CNN) model inspired by the architecture described as Model-1 in \cite{bacioiu_automated_2019}. This model is used both for feature extraction in the hybrid quantum-classical pipeline and as a classifier for end-to-end image-based defect detection, depending on whether the final fully connected (FC) layer is included. If the final FC layer is present, the model performs classification; else, the output from the preceding layer is used as a feature vector input to the hybrid quantum-classical pipeline. The CNN comprises a series of convolutional layers with max-pooling and activation functions, designed to progressively abstract spatial features from the input images. These convolutional layers learn hierarchical representations of weld characteristics such as texture, edge continuity, and defect patterns. The extracted spatial features are then flattened and passed through one or more dense layers. For the classical classification pipeline, we append a final fully connected layer followed by a softmax activation function as shown in Figure~\ref{fig:cnn}. The number of outputs in this layer determines the classification tasks: two for binary classification (good vs. defective welds) and three or more for multiclass classification (good weld, contamination, and lack of fusion). The softmax function transforms the raw outputs into normalized class probabilities, allowing the model to assign confidence scores to each class. The network is trained using categorical cross-entropy loss and optimized via backpropagation using the Adam optimizer for different feature sizes.

This classical CNN serves as a strong benchmark due to its proven performance in previous studies and industrial applications. Moreover, it forms the backbone of our hybrid models by supplying robust feature vectors to the quantum circuits. Thus, its role is central not only in the classical workflow but also as a bridge between raw image data and quantum processing in the hybrid architectures.

\subsection{VQLS-Enhanced QSVM}\label{subsec:vqls}

The VQLS-enhanced QSVM approach extends the classical support vector machine (SVM) into the quantum domain by leveraging a quantum kernel and solving the associated optimization problem using a Variational Quantum Linear Solver (VQLS). The pipeline of this algorithm is illustrated in Figure~\ref{fig:vqls}. This hybrid model aims to classify images based on quantum-encoded features while maintaining compatibility with Noisy Intermediate-Scale Quantum (NISQ) hardware. The pseudocode for this method is given in Algorithm~\ref{alg:vqlsalg}. The main steps are explained in detail as follows:

\begin{figure}[H]
    \centering
    \includegraphics[width=0.9\textwidth]{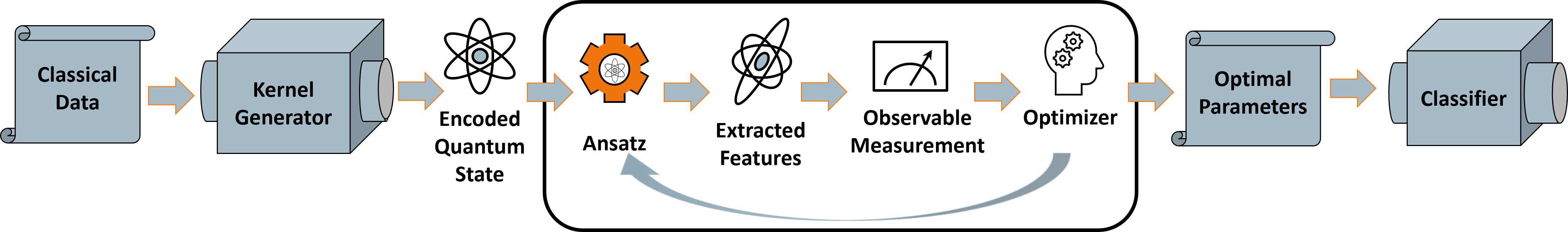}
    \caption{Pipeline for VQLS-enhanced QSVM algorithm.\label{fig:vqls}}
\end{figure}

\subsubsection{Feature Normalization}
To ensure stable performance during quantum processing, the CNN-extracted feature vectors are first normalized. Normalization is necessary because these vectors are used to prepare quantum states, which require unit norm (i.e., $\norm{\mathbf{x}}=1$). Let $\mathbf{x}_i \in \mathbb{R}^d$ be the feature vector for sample $i$, the normalized vector is given by: $\tilde{\mathbf{x}}_i = \frac{\mathbf{x}_i}{\norm{\mathbf{x}_i}}$. These normalized vectors are then encoded into quantum states $|\tilde{x}_i\rangle$.

\subsubsection{Quantum Kernel Construction}
The quantum kernel matrix $K \in \mathbb{R}^{n \times n}$ is constructed using the inner products between quantum-encoded data points:

\begin{equation}
K_{ij} = |\langle \tilde{x}_i | \tilde{x}_j \rangle|^2
\end{equation}

This kernel effectively maps classical features into a higher-dimensional Hilbert space where linear decision boundaries become more discriminative.

\subsubsection{Linear System Formulation}
For a soft-margin linear SVM, the optimal separating hyperplane can be obtained by solving a linear system \cite{10.5555/3104482.3104606}:

\begin{equation}
M \boldsymbol{\alpha} = \mathbf{b}
\end{equation}

where $M = K + \lambda I$, $\boldsymbol{\alpha}$ are the dual coefficients, $\lambda$ is a regularization parameter, and $\mathbf{b}$ contains the labels in the dual formulation.

This equation is translated into the quantum domain as:

\begin{equation}
\hat{M} |\alpha\rangle = |b\rangle,
\end{equation}

where $\hat{M}$ is a Hermitian matrix obtained via \textbf{Pauli decomposition} of the classical matrix $M$, $\hat{M} = \sum_j c_j P_j$ and $|\alpha\rangle$ is the quantum state approximating the solution.

\subsubsection{Hardware-Efficient Ansatz Construction}
\begin{figure}[H]
    \centering
    \includegraphics[width=0.5\textwidth]{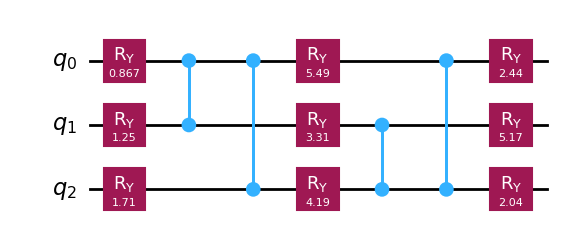}
    \caption{Quantum circuit for hardware-efficient ansatz.\label{fig:vqls_ansatz}}
\end{figure}

To approximate the solution $|\alpha\rangle$, we construct a parameterized quantum circuit (ansatz) using rotational $R_y(\theta)$ and entangling controlled-$R_z(\phi)$ gates as shown in Figure~\ref{fig:vqls_ansatz}. The parameters are initialized randomly and optimized during training.

\subsubsection{Cost Function via Hadamard Test}
A global cost function quantifies how close the current ansatz state $| \alpha(\boldsymbol{\theta}) \rangle$ is to the true solution $|b\rangle$ of the linear system $\hat{M}|\alpha(\boldsymbol{\theta})\rangle=|b\rangle$, where $\hat{M}$ is the Hermitian matrix representing the linear system operator. The cost function $C(\boldsymbol{\theta})$ is given by:

\begin{equation}
C(\boldsymbol{\theta}) = 1 - \frac{|\langle b | \hat{M} | \alpha(\boldsymbol{\theta}) \rangle|^2}{\langle \alpha(\boldsymbol{\theta}) | \hat{M}^2 | \alpha(\boldsymbol{\theta}) \rangle}
\end{equation}

The numerator $(|\langle b | \hat{M} | \alpha(\boldsymbol{\theta}) \rangle|^2)$ captures the overlap between the trial solution $\hat{M}|\alpha(\boldsymbol{\theta})\rangle$ and the true target vector $|b\rangle$, evaluated using Hadamard tests, while the denominator $(\langle \alpha(\boldsymbol{\theta}) | \hat{M}^2 | \alpha(\boldsymbol{\theta}) \rangle)$ normalizes the overlap. Optimization of the variational parameter $\boldsymbol{\theta}$ to minimize $C(\boldsymbol{\theta})$ is performed using the classical COBYLA optimizer \cite{Powell1994ADS}.

\subsubsection{Solution State Preparation}
Upon convergence, the final ansatz circuit represents the state, from which we measure the statevector to obtain the amplitude distribution:

\begin{equation}
|\alpha^*\rangle = \sum_i c_i |i\rangle
\end{equation}

These coefficients $c_i$ are used to construct the solution vector corresponding to the decision function. To extract meaningful classification parameters, we rescale the amplitudes using a linear regression model trained on the classical dual SVM form, yielding the final model parameters:

\begin{equation}
\boldsymbol{\theta} = A \boldsymbol{\theta}' + \mathbf{b}, \quad \text{bias} = b_0
\end{equation}

\subsubsection{Decision Function}
The final classifier uses the quantum kernel $K(x_i, x)$ and trained parameters to evaluate new inputs $x$ using:

\begin{equation}
f(x) = \sum_i \alpha_i K(x_i, x) + b,
\end{equation}

where $x_i$ represents the support vector data points, $\alpha_i$ are the corresponding optimized coefficients and $b$ is the bias.

For multiclass classification, we adopt a one-vs-rest (OvR) approach: a separate binary classifier is trained for each class against all others. During inference, the classifier with the highest decision value determines the predicted class. This strategy allows the VQLS-enhanced QSVM framework to be effectively extended beyond binary tasks.

\begin{algorithm}[H]
\caption{VQLS-enhanced QSVM}
\label{alg:vqlsalg}
\begin{algorithmic}
\State \textbf{Input:} Feature samples $\mathbf{X}_{\text{train}} = \{\mathbf{x}_1, \cdots, \mathbf{x}_N\}$, labels $\mathbf{y}_{\text{train}} = \{y_1, \cdots, y_N\}$
\State \textbf{Output:} Optimal parameters $\boldsymbol{\alpha}_{\text{opt}}$
\State Normalize $\mathbf{X}_{\text{train}} \rightarrow \hat{\mathbf{X}}_{\text{train}}$
\State Construct kernel matrix $K$ using quantum kernel estimation
\State Pauli decomposition of $K$ into $\sum_{\ell=0}^{N} c_\ell A_\ell$ with coefficients $c_\ell$ and pauli strings $A_\ell$
\State Initialize $i \gets 0$, cost $C \gets 1$, threshold $\epsilon \gets 0.01$, max iterations $T_{\text{max}} \gets 300$
\State Randomly initialize parameters $\boldsymbol{\alpha}^{(0)}$
\While{$C > \epsilon$ \textbf{and} $i < T_{\text{max}}$}
    \State $sum_1 \gets 0$
    \For{$A_m$ in $\{A_1, \cdots, A_N\}$}
        \For{$A_n$ in $\{A_1, \cdots, A_N\}$}
            \State Construct Circuit-1 to compute $\langle 0|V(\alpha)^\dagger A_m^\dagger A_n V(\alpha)|0\rangle$ using Hadamard test
            \State Execute with 10,000 shots, measure ancilla qubit with overlap results $\rightarrow$ get expectation value
            \State $sum_1 \gets sum_1 + c_m^* c_n \cdot$ measured value
        \EndFor
    \EndFor
    \State $sum_2 \gets 0$
    \For{$A_m$ in $\{A_1, \cdots, A_N\}$}
        \For{$A_n$ in $\{A_1, \cdots, A_N\}$}
        \State Construct Circuit-2 for overlap of $|b \rangle$ with ansatz after $A_n$; Circuit-3 for overlap of ansatz after $A_m^\dagger$ with $\lvert b \rangle$
            \State Circuit-2 to compute $\langle 0|U^\dagger A_n V(\alpha)|0\rangle$
            \State Execute, measure, store result
            \State Circuit-3 for $\langle 0|V(\alpha)^\dagger A_m^\dagger U|0\rangle$
            \State Execute, measure, store result
            \State $sum_2 \gets sum_2 + c_m^* c_n$ product of both results
        \EndFor
    \EndFor
    \State Compute $C \gets 1 - \frac{sum_1}{sum_2}$
    \State $i \gets i + 1$
    \State Update $\boldsymbol{\alpha}^{(i)}$ using COBYLA optimizer
\EndWhile
\State \Return $\boldsymbol{\alpha}_{\text{opt}}$
\State Measure final ansatz state $\boldsymbol{\alpha}_{\text{opt}}$ to extract amplitudes $\theta'$
\State Apply linear regression: $\theta = A \cdot \theta' + b$
\State Define decision function: $f(x) = \sum_i \alpha_i K(x_i, x) + \text{bias}$
\State \Return $\boldsymbol{\alpha}_{\text{opt}}$, $f(x)$
\end{algorithmic}
\end{algorithm}

\subsection{Variational Quantum Circuit (VQC)-Based Classifier}
\label{subsec:vqc}

The second quantum pipeline in our study is inspired by the hybrid transfer learning approach proposed by Geng et al. in \cite{geng_hybrid_2025}. Unlike the VQLS-enhanced QSVM, which solves a quantum linear system for classification, this model uses a variational quantum circuit trained via gradient descent. The classical component first reduces the input feature dimension, while the quantum circuit provides a trainable embedding to enhance classification performance. The pseudocode for this method is given in Algorithm~\ref{alg:VQC}, the pipeline is illustrated in Figure~\ref{fig:vqc} and the key steps are outlined below:

\begin{figure}[H]
    \centering
    \includegraphics[width=0.9\textwidth]{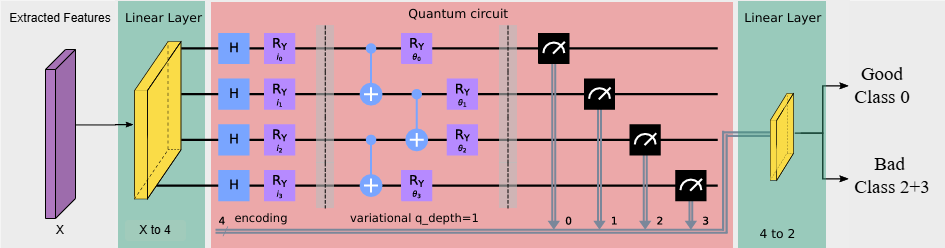}
    \caption{Pipeline for VQC-based classifier algorithm for binary classification.\label{fig:vqc}}
\end{figure}

\subsubsection{Dimensionality Reduction}
Given a classical feature vector $\mathbf{x} \in \mathbb{R}^d$ extracted from a CNN, we first reduce its dimension to match the number of available qubits. A linear transformation is applied:

\begin{equation}
    \mathbf{x}' = W \mathbf{x} + \mathbf{b}, \quad \mathbf{x}' \in \mathbb{R}^4
\end{equation}

This projection retains critical semantic information while enabling feasible quantum encoding on near-term devices.

\subsubsection{Quantum Embedding}
Each element $x_i'$ of the reduced feature vector is encoded into the state of a qubit using a parameterized single-qubit rotation gate:

\begin{equation}
    |\psi_{\text{init}}\rangle = \bigotimes_{i=1}^{4}  R_y(x_i') H |0\rangle
\end{equation}

\subsubsection{Variational Quantum Circuit (VQC)}
The quantum processing is carried out by a hardware-efficient ansatz, constructed from trainable $R_y(\theta_i)$ gates and entangling $CX$ gates. The unitary transformation $U(\theta)$ prepares the trial quantum state:

\begin{equation}
    |\psi(\theta)\rangle = U(\theta) |\psi_{\text{init}}\rangle
\end{equation}

We use a single variational layer, which strikes a good balance between maintaining accuracy and minimizing the number of circuit executions, helping to manage noise and errors in NISQ devices.

\subsubsection{Measurement}
Each qubit is measured in the Pauli-Z basis to yield a 4-dimensional output vector:

\begin{equation}
    z_i = \langle \psi(\theta)| Z_i | \psi(\theta) \rangle, \quad i = 1,\dots,4
\end{equation}

\begin{equation}
    \mathbf{z} = [z_1, z_2, z_3, z_4] \in \mathbb{R}^4
\end{equation}

\subsubsection{Post-Quantum Processing}
The measured quantum outputs are passed through a final classical linear layer followed by a softmax function for binary classification:

\begin{equation}
    \hat{y} = \text{Softmax}(W_z \mathbf{z} + \mathbf{b}_z)
\end{equation}

\subsubsection{Training}
The model is trained using cross-entropy loss and the Adam optimizer. Gradients are computed using the parameter-shift rule. All experiments are performed on the \texttt{PennyLane default.qubit} simulator, running locally on a CUDA-enabled GPU.

For multiclass classification, the VQC-based classifier model adjusts the number of outputs in the final classical linear layer to match the number of classes two for binary, three or more for multiclass tasks. The softmax activation then produces class probabilities accordingly, enabling seamless extension from binary to multiclass classification.

\begin{algorithm}[H]
\caption{VQC-based Classifier}
\label{alg:VQC}
\begin{algorithmic}
\State \textbf{Input:} Feature vectors $X_{\text{train}} = \{\mathbf{x}_1, \cdots, \mathbf{x}_N\}$ and labels $y_{\text{train}}$
\State \textbf{Output:} Trained VQC parameters $\theta^{*}$ and classifier weights
\For{each input $\mathbf{x}_i$}
    \State Reduce to 4D: $\mathbf{x}_i' = W \mathbf{x}_i + \mathbf{b}$
    \State Encode $\mathbf{x}_i'$ using $R_y$ gates: $\ket{\psi_{\text{init}}}$
    \State Apply VQC: $\ket{\psi(\theta)} = U(\theta)\ket{\psi_{\text{init}}}$
    \State Measure $\langle Z_i \rangle$ on each qubit to obtain $\mathbf{z}$
    \State Compute prediction: $\hat{y}_i = \text{Softmax}(W_z \mathbf{z} + \mathbf{b}_z)$
\EndFor
\State Compute loss: $\mathcal{L} = \text{CrossEntropy}(\hat{y}, y)$
\State Compute gradients (parameter-shift rule or backpropagation)
\State Update parameters $\theta$, $W_z$, $\mathbf{b}_z$ using Adam
\State \Return Optimal $\theta^*$ and trained classifier
\end{algorithmic}
\end{algorithm}

\section{Dataset}\label{sec:data}

The Kaggle dataset used in this study originates from an industrial setting, where high-resolution image sequences of Tungsten Inert Gas (TIG) welding processes were captured using a high dynamic range (HDR) camera system \cite{kaggle_tig_5083}. The primary objective of the data acquisition was to support non-destructive testing (NDT) of aluminium welds by enabling real-time defect detection through visual inspection without damaging the material or structure.

\begin{figure}[H]
    \centering
    \includegraphics[width=0.7\textwidth]{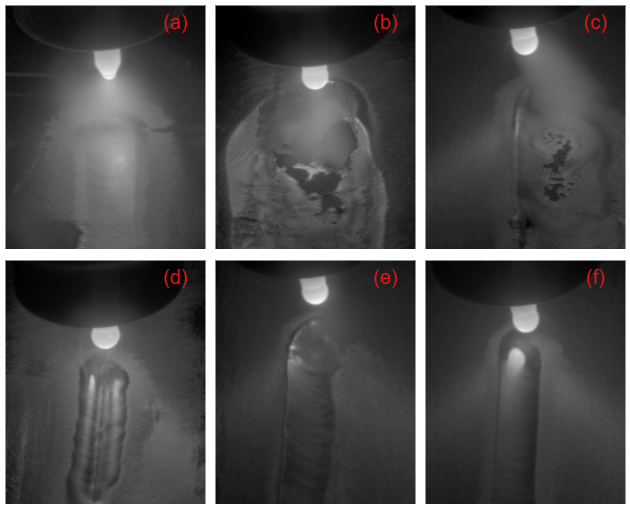}
    \caption{Six class categorization of welding defects: a) Good Weld, b) Burn through, c) Contamination, d) Lack of fusion, e) Misalignment, and f) Lack of penetration. These images are taken from Kaggle \cite{kaggle_tig_5083}.\label{fig:6class}}
\end{figure}

The original dataset contains approximately 30,000 grayscale images, recorded at a sampling rate of 55 frames per second. Each image has a native resolution of 800 $\times$ 974 pixels, providing fine spatial details essential to recognize subtle weld imperfections. A team of welding experts manually annotated the dataset, categorizing the images into six classes corresponding to various types of welding defects as shown in Figure~\ref{fig:6class}.

\begin{figure}[H]
    \centering
    \includegraphics[width=0.75\textwidth]{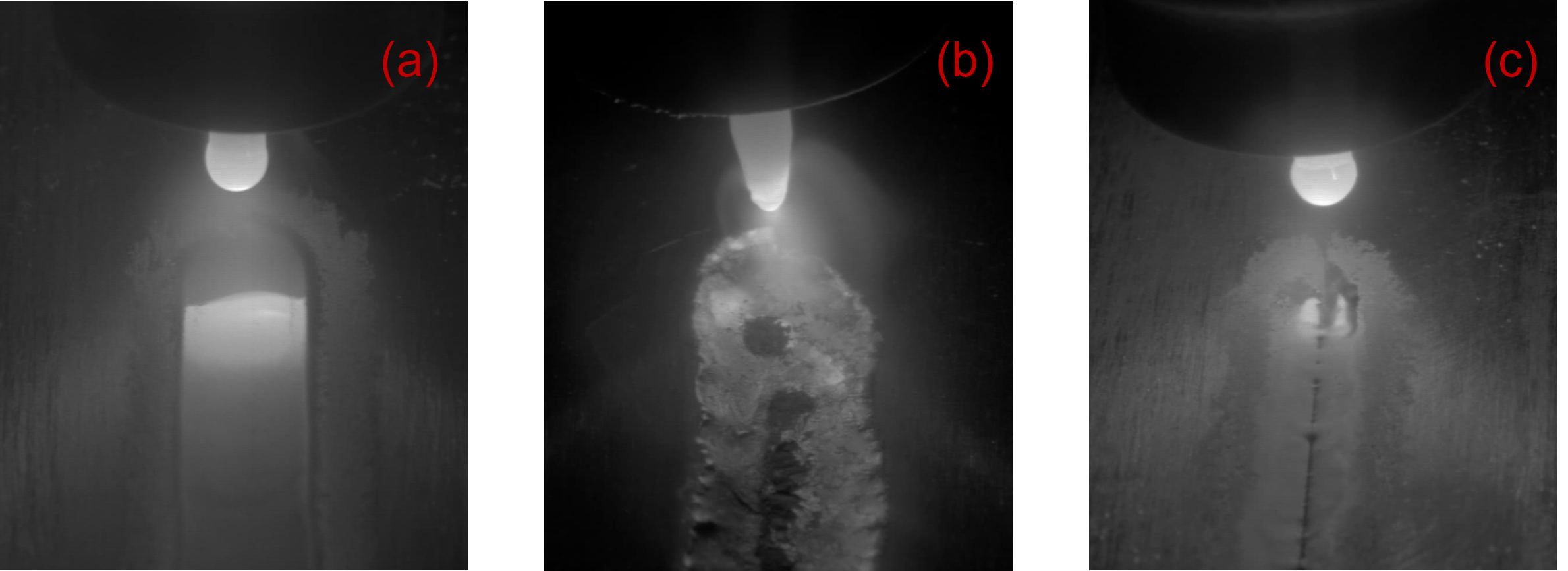}
    \caption{Three class categorization of welding defects: a) Good Weld, b) Contamination, and c) Lack of fusion.\label{fig:3class}}
\end{figure}

For this study, we curated a representative subset of this dataset, focusing on a three-class classification task. The selected classes include: Class 0 - Good weld, Class 2 - Contamination, and Class 3 - Lack of fusion. These three categories capture a meaningful distinction between good welds and two common, visually identifiable defect types, facilitating robust model training and evaluation, as seen in Figure~\ref{fig:3class}.

The images were spatially downsampled to a resolution of $400 \times 487$ pixels to reduce computational load while preserving defect-relevant features. We used a total of 1,100 images in our study, allocating 700 for training and validation and 400 (100 of good welds, 150 each for defect classes 2 and 3) for testing. The dataset was balanced across the selected classes to ensure fair evaluation during both the training and testing phases. This carefully designed subset serves as a reliable benchmark for assessing the performance of both classical and hybrid quantum-classical machine learning models in industrial visual inspection scenarios.

\section{Hyperparameter Tuning} \label{sec:hypertuning}

To get the best performance from our hybrid quantum-classical classifier, we tuned several key hyperparameters using the Ray Tune Python package \cite{liaw2018tune}, which offers scalable and efficient optimization. Since quantum circuits are sensitive to noise and training is expensive, careful tuning was necessary to ensure both stability and accuracy.

\subsection{Tuned Hyperparameters}

The following parameters were explored:

\begin{itemize}
    \item Batch size - Tested values: 4, 8, 12, 16. Smaller batch sizes provided more frequent updates and some regularization, while larger ones helped with more stable gradient estimates.
    
    \item Feature Size - We tried feature vectors of sizes \cite{yi_variational_2023} $7, 15, 31, 63, \text{and}~127$. These come from the CNN feature extractor before quantum encoding. Larger feature sizes can preserve more information but increase the complexity of the quantum circuit.
    
    \item Number of Qubits - Set based on the reduced feature size, with values 4, 7, 12, and 15. Each qubit represents one encoded feature.
    
    \item Quantum Depth - We fixed the variational circuit depth to 1 as discussed by Geng in \cite{geng_application_2024}. Deeper circuits could improve model capacity, but are time-consuming and more prone to noise, especially on current hardware.
\end{itemize}

\subsection{Search Method - Population-Based Training (PBT)}

We used PBT from the Ray Tune package to optimize hyperparameters efficiently. This method trains multiple models in parallel, each with different hyperparameters. After every perturbation interval, models are compared. The better ones are kept (``exploited''), while others are replaced or slightly modified (``explored'').

\begin{itemize}
    \item Perturbation interval - every 5 epochs
    \item Resample probability - 0.3 (i.e., 30\% of models explore new values)
    \item Objective - Minimize training loss
\end{itemize}

PBT works well for our setup because it balances trying new settings and sticking with those that work. It's also adaptive as it tunes hyperparameters during training, which is useful in a noisy, hybrid quantum setup.

\subsection{Training Setup}

All hyperparameter optimization experiments used the \texttt{PennyLane} \texttt{default.qubit} simulator, running on a local GPU machine. We used Adam optimizer with cross-entropy loss, and trained each configuration for up to 50 epochs. Random seeds and model checkpoints were used to improve reproducibility.

\begin{figure}[H]
    \centering
    \includegraphics[width=0.75\textwidth]{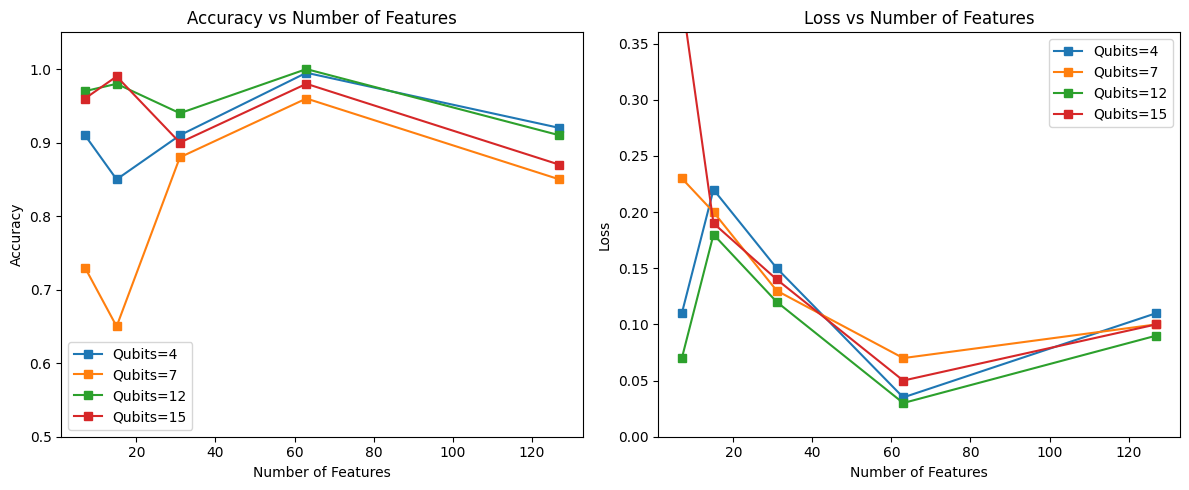}
    \caption{Hyperparameter tuning for the VQC-based classifier over feature sizes (7, 15, 31, 63, 127) and qubit configurations (4, 7, 12, 15), with a fixed batch size of 16.\label{fig:hyperpara}}
\end{figure}

\subsection{Best Configuration}
As shown in Figure~\ref{fig:hyperpara}, the hyperparameter search reveals that performance improves as the feature size increases from 7 to 63, after which gains slightly degrade, likely due to overfitting or encoding noise. Notably, the 12-qubit configuration achieves the highest accuracy and lowest loss at feature size 63; the results also show that 4 qubits produce comparable loss and accuracy values, with only a marginal performance difference. Considering the best trade-off between model complexity, qubit limitations of near-term NISQ hardware, and generalization capability, the optimal configuration was identified as a feature size of 63, 4 qubits, and a batch size of 16. This batch size was selected based on prior tuning, as it provided a stable balance between gradient estimation and convergence speed. This configuration was used for all subsequent benchmarking experiments.

\section{Results} \label{sec:results}

We evaluated the performance of three models classical CNN, VQLS-enhanced QSVM, and VQC-based classifier on binary (good vs. defective welds) and multiclass (three-class: good weld, contamination, and lack of fusion) classification tasks at feature size 63. The classical CNN, implemented following the architecture in~\cite{bacioiu_automated_2019}, was trained using the Adam optimizer with a learning rate of $1 \cdot 10^{-4}$ and batch size of 10. Its performance was measured using accuracy and loss metrics on a fixed test set of 400 images across different feature sizes $(7, 15, 31, 63, \text{and } 127)$. The VQLS-enhanced QSVM used feature vectors extracted from this CNN model, which were amplitude-encoded into quantum states to compute a quantum kernel matrix. The associated linear system was then solved variationally using a hardware-efficient ansatz with three qubits, optimized by the COBYLA algorithm to minimize the cost function described in the previous section. These experiments were performed on the \texttt{qiskit aer simulator} using 10,000 shots per circuit. For each feature size, training continued until the cost function converged or a maximum number of iterations was reached, and model performance was again evaluated on the same test set. The VQC-based classifier followed a hybrid design where the CNN-extracted features were first linearly reduced in dimension, then embedded into a four-qubit quantum circuit via $H$ and $R_y(x_i)$ rotations. A shallow variational ansatz processed these inputs, and Pauli-Z expectation values were passed through a final classical linear-softmax layer for prediction. Training was conducted for over 50 epochs with batch size 16 on the \texttt{PennyLane} \texttt{default.qubit} simulator using the Adam optimizer, cross-entropy loss, and the parameter-shift rule for gradient computation. The number of qubits was fixed at four, as determined through hyperparameter tuning discussed in the previous section.

\subsection{Binary Classification}

\subsubsection{Classical CNN}

The classical CNN model was evaluated on the binary classification task (good vs. defective welds) for different feature sizes. As shown in Table~\ref{tab:cnn_binary_results}, the model achieved perfect accuracy in all feature sizes with a steady decrease in loss as the feature size increased.

\begin{table}[H]
\caption{Binary classification results using the classical CNN model on the test set (400 images).\label{tab:cnn_binary_results}}
\begin{tabularx}{\textwidth}{CCCCCC}
\toprule
\textbf{Feature size} & \textbf{7} & \textbf{15} & \textbf{31} & \textbf{63} & \textbf{127} \\
\midrule
Accuracy & 1.000 & 1.000 & 1.000 & 1.000 & 1.000 \\
Loss     & 0.0204 & 0.0066 & 0.0145 & 0.0020 & 0.0003 \\
\bottomrule
\end{tabularx}
\end{table}

\subsubsection{VQLS-Enhanced QSVM}

Table~\ref{tab:vqls-binary} shows the binary classification results for the VQLS-enhanced QSVM. Accuracy improved with larger feature sizes, peaking at 63 features with 96.8\% accuracy and a loss of 0.2011. Performance dropped slightly at 127 features, likely due to circuit noise and overfitting. Training was performed until convergence, with the number of iterations varying depending on input properties such as the linear system's condition number $(\kappa)$.

\begin{table}[H]
\caption{Binary classification results using the VQLS-enhanced QSVM model on the test set (400 images).\label{tab:vqls-binary}}
\begin{tabularx}{\textwidth}{CCCCCC}
\toprule
\textbf{Feature size} & \textbf{7} & \textbf{15} & \textbf{31} & \textbf{63} & \textbf{127} \\
\midrule
Accuracy & 0.753 & 0.856 & 0.871 & 0.968 & 0.948 \\
Loss     & 0.454 & 0.395 & 0.341 & 0.201 & 0.238 \\
\bottomrule
\end{tabularx}
\end{table}

\subsubsection{VQC-Based Classifier}

The VQC-based classifier results in Table~\ref{tab:vqc-binary} demonstrated strong performance, with accuracy increasing alongside feature size and loss decreasing correspondingly. The best results were observed at a feature size of 63. Beyond this, accuracy slightly declined, likely due to specific feature characteristics impacting the model's effectiveness.

\begin{table}[H]
\caption{Binary classification results using the VQC-based classifier model on the test set (400 images).\label{tab:vqc-binary}}
\begin{tabularx}{\textwidth}{CCCCCC}
\toprule
\textbf{Feature size} & \textbf{7} & \textbf{15} & \textbf{31} & \textbf{63} & \textbf{127} \\
\midrule
Accuracy & 0.915 & 0.876 & 0.925 & 0.997 & 0.951 \\
Loss     & 0.126 & 0.235 & 0.164 & 0.042 & 0.113 \\
\bottomrule
\end{tabularx}
\end{table}

\begin{figure}[H]
    \centering
    \subfloat[\centering]{\includegraphics[width=0.45\textwidth]{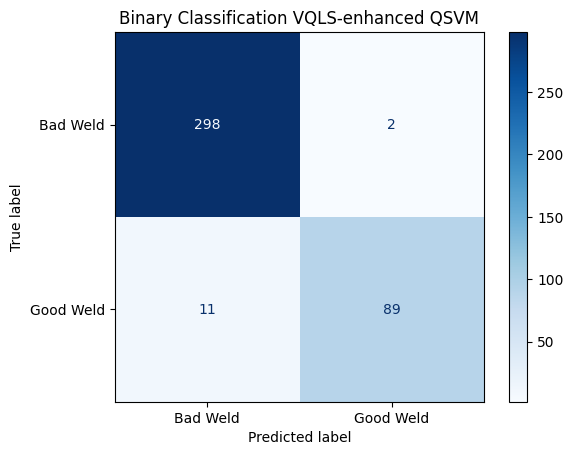}\label{fig:vqls_binary}}
    \hfill
    \subfloat[\centering]{\includegraphics[width=0.45\textwidth]{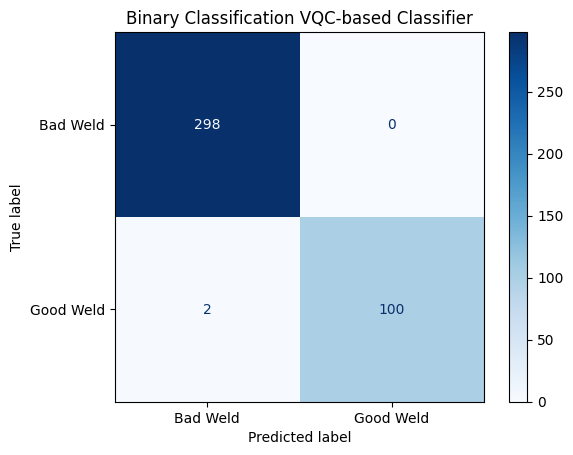}\label{fig:vqc_binary}}
    \caption{Confusion matrices for binary classification at feature size 63: (\textbf{a}) VQLS-enhanced QSVM, (\textbf{b}) VQC-based classifier.\label{fig:binary_cm}}
\end{figure}

The confusion matrices for the binary classification task at feature size 63 are shown in Figure~\ref{fig:binary_cm}. The VQLS-enhanced QSVM exhibits a small number of misclassifications, whereas the VQC-based classifier achieves near-perfect separation, with almost all samples correctly assigned to their respective classes, confirming the accuracy values reported in the tables above.

\subsection{Multiclass Classification}

\subsubsection{Classical CNN}

For the three-class classification task (good weld, contamination, lack of fusion), the classical CNN achieves high accuracy for all feature sizes (see Table~\ref{tab:cnn_multiclass_results}). The lowest loss is observed at a feature size of 63, suggesting that this size may provide an optimal trade-off between model complexity and generalization performance.

\begin{table}[H]
\caption{Multiclass classification results using the classical CNN model on the test set (400 images).\label{tab:cnn_multiclass_results}}
\begin{tabularx}{\textwidth}{CCCCCC}
\toprule
\textbf{Feature size} & \textbf{7} & \textbf{15} & \textbf{31} & \textbf{63} & \textbf{127} \\
\midrule
Accuracy & 1.000 & 0.9925 & 0.9975 & 1.000 & 0.9975 \\
Loss     & 0.0326 & 0.0288 & 0.0056 & 0.0011 & 0.0309 \\
\bottomrule
\end{tabularx}
\end{table}

\subsubsection{VQLS-Enhanced QSVM Multiclass}

Table~\ref{tab:vqls-multi} presents the results from three-class classification for the VQLS-enhanced QSVM. Similar to the binary case, three-class classification accuracy improved with larger feature sizes. The model achieved its highest accuracy of $92.4\%$ at 63 features with the lowest loss. However, overall accuracy dropped compared to the binary case, due to an increase in circuit complexity and kernel evaluation demands. As with the binary task, increasing the feature size beyond 63 did not yield further gains, supporting the conclusion that this size provides the best balance between information richness and circuit feasibility. 

\begin{table}[H]
\caption{Multiclass classification results using the VQLS-enhanced QSVM model on the test set (400 images).\label{tab:vqls-multi}}
\begin{tabularx}{\textwidth}{CCCCCC}
\toprule
\textbf{Feature size} & \textbf{7} & \textbf{15} & \textbf{31} & \textbf{63} & \textbf{127} \\
\midrule
Accuracy & 0.625 & 0.700 & 0.772 & 0.924 & 0.884 \\
Loss     & 0.684 & 0.482 & 0.437 & 0.359 & 0.384 \\
\bottomrule
\end{tabularx}
\end{table}

\subsubsection{VQC-Based Classifier Multiclass}

The VQC-based classifier exhibited strong multiclass performance with an accuracy of 98.9\% at 63 features (Table~\ref{tab:vqc-multi}). Lower feature sizes led to reduced accuracy caused by insufficient information, while larger feature sizes introduced marginal performance drops, possibly due to data characteristics and overfitting effects.

\begin{table}[H]
\caption{Multiclass classification results using the VQC-based classifier model on the test set (400 images).\label{tab:vqc-multi}}
\begin{tabularx}{\textwidth}{CCCCCC}
\toprule
\textbf{Feature size} & \textbf{7} & \textbf{15} & \textbf{31} & \textbf{63} & \textbf{127} \\
\midrule
Accuracy & 0.890 & 0.892 & 0.942 & 0.989 & 0.946 \\
Loss     & 0.203 & 0.209 & 0.126 & 0.107 & 0.115 \\
\bottomrule
\end{tabularx}
\end{table}

\begin{figure}[H]
    \centering
    \subfloat[\centering]{\includegraphics[width=0.45\textwidth]{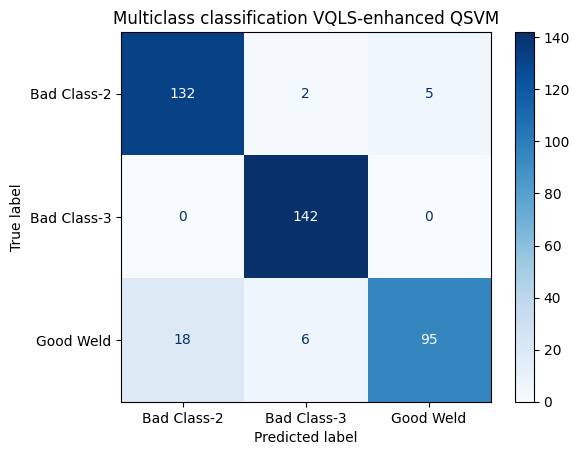}\label{fig:vqls_multi}}
    \hfill
    \subfloat[\centering]{\includegraphics[width=0.45\textwidth]{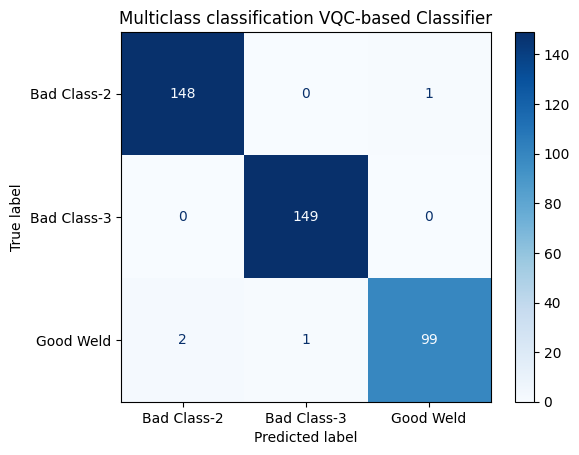}\label{fig:vqc_multi}}
    \caption{Confusion matrices for multiclass (3-class) classification at feature size 63: (\textbf{a}) VQLS-enhanced QSVM, (\textbf{b}) VQC-based classifier.\label{fig:multi_cm}}
\end{figure}

The confusion matrices for the three-class task at feature size 63 are presented in Figure~\ref{fig:multi_cm}. The VQLS-enhanced QSVM shows a notable increase in misclassifications compared to the binary case, reflecting the added difficulty of distinguishing three classes with a kernel-based approach. In contrast, the VQC-based classifier retains the performance, demonstrating its strong ability to separate all three weld categories even in the complex multiclass scenario.

When comparing the three models classical CNN, VQLS-enhanced QSVM, and the VQC-based classifier on identical binary and multiclass classification tasks, the classical CNN consistently maintained perfect accuracy across all feature sizes in both scenarios. This performance reflects its reliability and strong generalization capabilities for image classification problems. The VQC-based classifier achieved near-perfect accuracy, on par with the classical CNN, particularly at a feature size of 63 in both binary and multiclass tasks. Its shallow variational quantum circuit, combined with classical preprocessing and post-processing layers, enabled efficient learning and inference. This architecture proved well-suited for near-term quantum (NISQ) devices, offering a good trade-off between quantum resource usage and performance. The VQLS-enhanced QSVM also delivered good accuracy in the binary classification task, with its best results observed at a feature size of 63. However, the training process was significantly more time-consuming due to the complexity of the variational quantum linear solver and the repeated circuit evaluations required for kernel matrix construction and cost minimization. These evaluations are performed iteratively until convergence, and the number of iterations can vary based on the input data properties such as the condition number $(\kappa)$ of the linear system, which directly impacts computational cost. Its performance declined in the multiclass scenario, primarily due to the increased circuit complexity and the higher dimensionality of quantum kernel evaluations required for distinguishing three classes. 

\begin{figure}[H]
    \centering
    \includegraphics[width=0.5\textwidth]{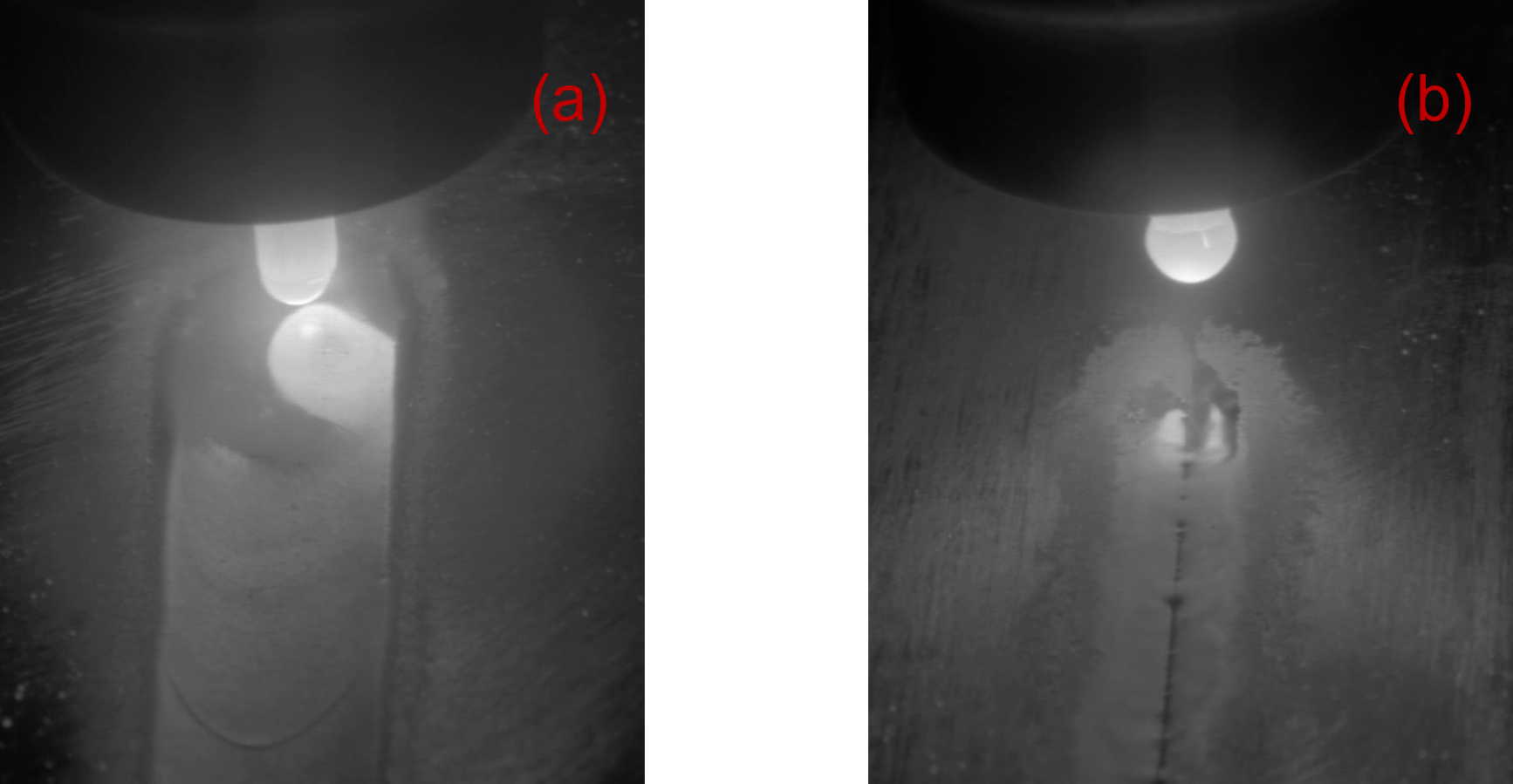}
    \caption{Visual cues influencing classification results. a) Good weld with reflection head resembling contamination, and b) Lack of Fusion defect with dark line in the center aiding detection.\label{fig:data}}
\end{figure}

Some of the observed variations in model performance across feature sizes can be due to the intrinsic characteristics of the dataset. Reflections from the weld head occasionally led to misclassifications, especially in the multiclass task, where some good welds visually resembled contamination (Class 2). However, contamination defects were typically well-localized and visually distinct, allowing all models to detect them reliably. In the case of fusion defects (Class 3), the presence of a dark line in the weld center served as a critical visual cue (as shown in Figure~\ref{fig:data}), which helped both the VQC and CNN models maintain high classification performance.

\section{Conclusions}

In this study, we investigated the classification of welding images using a hybrid classical-quantum algorithm on a high-resolution industrial dataset acquired from Tungsten Inert Gas (TIG) welding processes. We benchmarked three distinct pipelines: a classical convolutional neural network (CNN), a VQLS-enhanced quantum support vector machine (QSVM), and a hybrid variational quantum circuit (VQC)-based classifier. All three approaches use CNN-based feature extraction to reduce the high-dimensional image data into a manageable feature vector suitable for quantum encoding. This decision addresses the current hardware limitations of NISQ-era quantum devices. While the classical CNN achieved consistently high performance as expected, the hybrid quantum models also demonstrated competitive results. The VQC-based classifier achieved the best performance in binary classification, offering a strong balance between accuracy and quantum circuit depth. The VQLS-enhanced QSVM showed promising results as well, but faced practical limitations due to its circuit complexity and training overhead. Notably, the characteristics of the dataset also contributed to the high classification performance. Defects such as contamination and lack of fusion were visually distinct, which likely made the classification task more tractable across all models. Overall, our results indicate that hybrid quantum-classical architectures can already offer competitive performance on real-world industrial tasks such as weld defect detection. As quantum hardware advances, such approaches could become increasingly practical and scalable for applications in automated industrial inspection, which require high sensitivity like non-destructive testing. 


\section*{Funding}
This work was partially funded by the Federal Ministry for Economic Affairs and Climate Action through the EniQmA project (funding number 01MQ22007A).

\section*{Acknowledgments}
This work was supported by the Quantum Initiative Rhineland-Palatinate (QUIP) and the Interreg Upper Rhine project UpQuantVal. We also acknowledge the Fraunhofer-wide Capacity Building Campaign Quantum Now, which enabled us to perform experiments on both U.S. and EU-based IBM Quantum systems.

\bibliographystyle{unsrt} 
\bibliography{references}

\end{document}